\documentclass[letterpaper, 10 pt, conference]{ieeeconf}  

\IEEEoverridecommandlockouts                              

\pdfoutput=1

\usepackage{graphics} 
\usepackage{epsfig} 
\usepackage{times} 
\usepackage{amsmath} 

\DeclareMathOperator*{\argmax}{argmax}
\usepackage{amssymb}  

\usepackage{booktabs}
\usepackage{dblfloatfix}
\usepackage{graphicx} 
\usepackage{caption}

\let\labelindent\relax
\usepackage{enumitem}
\setlist[itemize,1]{leftmargin=\dimexpr 26pt-.2in}
\setlist[enumerate,1]{leftmargin=\dimexpr 26pt-.2in}
\usepackage{cuted} 
\usepackage{capt-of}

\usepackage{subcaption}
\usepackage{bbding}
\usepackage{color}
\usepackage{float} 
\usepackage{algorithm}
\usepackage[noEnd=True]{algpseudocodex}

\makeatletter
\let\NAT@parse\undefined
\makeatother
\usepackage{hyperref}  

\definecolor{MyDarkBlue}{rgb}{0,0.08,1}
\definecolor{MyDarkGreen}{rgb}{0.02,0.6,0.02}
\definecolor{MyDarkRed}{rgb}{0.8,0.02,0.02}
\definecolor{MyDarkOrange}{rgb}{0.40,0.2,0.02}
\definecolor{MyPurple}{RGB}{111,0,255}
\definecolor{MyRed}{rgb}{1.0,0.0,0.0}
\definecolor{MyGold}{rgb}{0.75,0.6,0.12}
\definecolor{MyDarkgray}{rgb}{0.66, 0.66, 0.66}

\renewcommand{\paragraph}[1]{\noindent {\bf #1}}

\newcommand{\kw}[1]{\textbf{#1}}

\newcommand{\methodtitle}{Embodied Uncertainty-Aware Object Segmentation}
\newcommand{\methodname}{embodied uncertainty-aware object segmentation}
\newcommand{\methodabbrev}{{\sc EOS}}

\newcommand{\uncsegproblemname}{{uncertainty-aware object instance segmentation}}
\newcommand{\uncsegproblemabbrev}{{\sc UncOS}} 

\newcommand{\uncsegmethodtitle}{Uncertainty-aware Object Segmentation Model}
\newcommand{\uncsegmethodname}{uncertainty-aware object segmentation model} 
\newcommand{\uncsegmethodabbrev}{{\sc UncOS}}

\newcommand{\bottomUpPt}{{\sc BUHighRecSeg}}
\newcommand{\bottomUpIm}{{\sc BUSeed}}
\newcommand{\topDown}{{\sc TDHighPrecSeg}}

\newcommand{\rgbd}{{\sc rgb-d}}
\newcommand{\rgb}{{\sc rgb}}
\newcommand{\sam}{{\sc sam}}
\newcommand{\etal}{\textit{et al.}}
\newcommand{\plus}{\texttt{+}}

\title{\LARGE \bf \methodtitle
}

\author{Xiaolin Fang, Leslie Pack Kaelbling, Tom\'as Lozano-P\'erez\\
MIT CSAIL\\
        {\tt\small \{xiaolinf,lpk,tlp\}@csail.mit.edu}
        \vspace{-20pt}}

\begin{document}

\maketitle
\thispagestyle{empty}
\pagestyle{empty}
\begin{strip} 
    \centering
    \includegraphics[trim={2.5cm 9cm 2cm 0.5cm},clip,width=\linewidth]{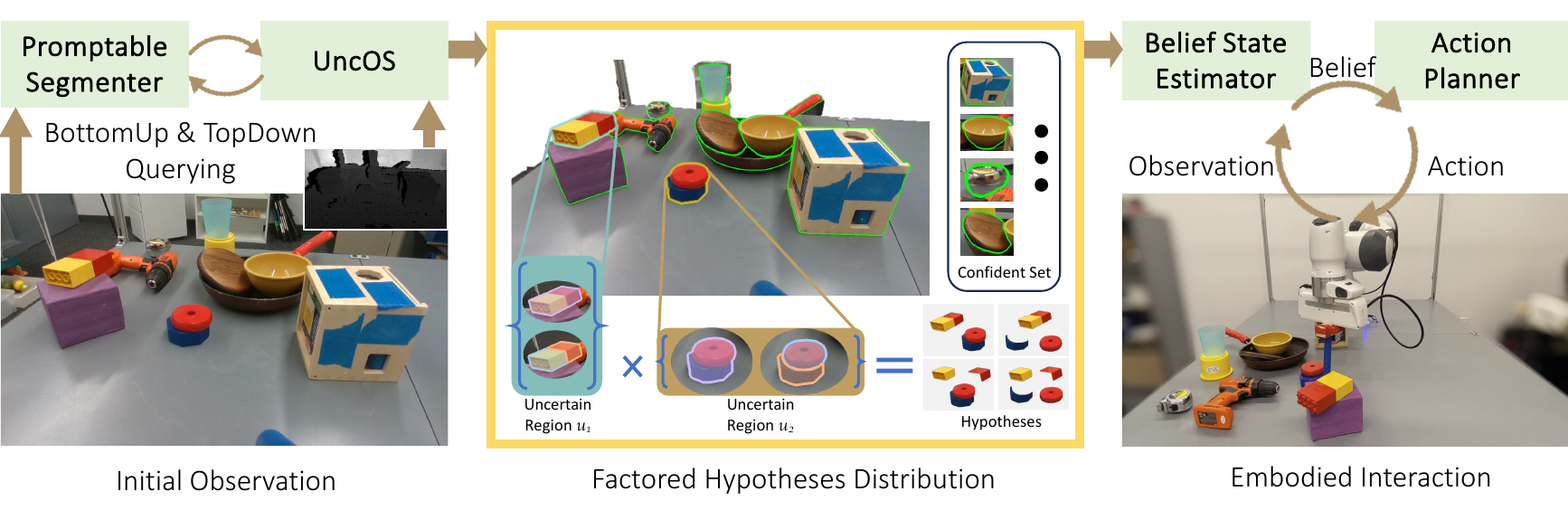}
    \captionof{figure}{Embodied segmentation with \uncsegmethodname~(\uncsegmethodabbrev{}) as a basis.
    \methodabbrev{} architecture:  an initial \rgbd{} image is repeatedly prompted by \uncsegmethodabbrev{} to obtain a region-based factored segmentation hypotheses distribution. Unambiguous regions are put into the confident set (outlined in green). Alternative hypotheses are proposed for each uncertain region. A distribution over segmentation hypotheses for the whole image is constructed by taking the Cartesian product of hypothesis distributions in each region. Such factored hypothesis distribution is used to initialize a 3D belief state that forms the basis for information-gathering action planning in embodied object segmentation.    
    }
    \vspace{-10pt}
    \label{fig:illustrative_belief}
\end{strip}

\begin{abstract}

We introduce {\em \uncsegproblemname{} (\uncsegproblemabbrev{})} 
and demonstrate its usefulness for embodied interactive segmentation. To deal with uncertainty in robot perception, we propose a method for generating a hypothesis distribution of object segmentation. We obtain a set of region-factored segmentation hypotheses together with confidence estimates by making multiple queries of large pre-trained models. This process can produce segmentation results that achieve state-of-the-art performance on unseen object segmentation problems. The output can also serve as input to a belief-driven process for selecting robot actions to perturb the scene to reduce ambiguity.  We demonstrate the effectiveness of this method in real-robot experiments. Website: \href{https://sites.google.com/view/embodied-uncertain-seg}{https://sites.google.com/view/embodied-uncertain-seg}.

\end{abstract}

\section{Introduction}

Our goal is to build long-horizon manipulation systems that can operate in environments that contain previously unknown objects.  A key step in such systems is segmenting images, either \rgb{} or \rgbd{}, into candidate objects to be manipulated.  This step is often called ``unknown object instance segmentation'' (UOIS) and a number of existing deep-learning models have been developed for this task~\cite{xie2021unseen, xiang2020learning, lu2022mean}.  However, the output from these models is inevitably imperfect, due to limitations in the model, for example, limited data or limited capacity, or to challenges in the images, for example, occlusion or lighting, or to fundamental ambiguity, for example, in a stack of toy blocks.  In the ``embodied'' manipulation setting, where we have a robot available, we can interact with the scene so as to obtain additional information, such as pushing some of the objects and tracking how they move.  Furthermore, with the advent of ``promptable'' segmentation models~\cite{kirillov2023segany}, we can also interact with the model to obtain additional information, such as obtaining multiple segmentations from different prompts.  In this paper, we explore both of these methodologies for improving segmentation results: multiple prompting of the segmentation model and active robot interaction with the objects.  In particular, we construct (by multiple prompting) a characterization of the uncertainty in the segmentation and use that representation to guide the physical interaction.

Image segmentation, in its most general form, is fundamentally underconstrained.  Is the bottle cap part of the bottle or a separate object?  Is the shirt part of the person?  In this paper, we limit ourselves to considering the segmentation of discrete rigid objects, where the answer to these questions is: if chunks of matter always move together rigidly, then they form a single object, and not otherwise.  It will typically be impossible to find this ground-truth segmentation from an image of a cluttered scene and, in general, it may not be necessary to find it so as to achieve a particular robotic manipulation goal. 

We define our task as that of {\em \uncsegproblemname{}}.  Given an image, a solution to the problem partitions a scene into disjoint regions and provides a single interpretation for each region of the scene that has sufficiently low uncertainty and provides multiple interpretations for each region with high uncertainty.  This is different from the classic instance segmentation task, where the objective is to deliver a single set of masks for the scene.  By explicitly characterizing this region-factored uncertainty, we hope to enable improved performance on downstream tasks, for example, improving the choice of actions to gather additional information to disambiguate the segmentation.

A crucial question in this approach is how to characterize the uncertainty in a proposed segmentation.  We develop an uncertainty estimation and hypotheses generation method based on multiple queries to large pre-trained, ``promptable''  models~\cite{kirillov2023segany, zhang2022dino}.  Within a region of the image, we issue random point prompts and use consistency of the returned masks as an indication of uncertainty.

Having obtained object hypotheses, with multiple candidate segmentation of uncertain regions, we use the robot to do targeted exploration aimed at reducing the uncertainty.  We use a maximal-uncertainty-reduction-driven action selection heuristic to lightly push a candidate object.  We build a state estimator to track and update the object hypotheses. The most likely segmentation hypothesis can be computed from the resulting belief state and we show that the state estimation leads to better action choices which ultimately leads to better maximum-likelihood segmentation hypotheses.



The key contributions of this work are:
\begin{itemize}
    \item \uncsegmethodabbrev{}: An active prompting strategy for combining promptable top-down and bottom-up pre-trained object instance segmentation methods to obtain a distribution over image-segmentation hypotheses;
    \item \methodabbrev{}: A method for converting this segmentation distribution into a distribution over world models and using that for selecting robot perturbation actions to disambiguate the segmentation. 
\end{itemize}
We demonstrate the effectiveness of the \uncsegmethodabbrev{} image-segmentation strategy first as a stand-alone method, showing that taking its maximum-likelihood hypothesis performs better than state-of-the-art UOIS methods.  Furthermore, we show that the hypothesis distribution produced by \uncsegmethodabbrev{} can be used by \methodabbrev{} to generate targeted physical interactions with the scene that gather information much more efficiently than less-informed alternatives.

\section{Related work}

This paper is related to previous work on unseen object instance segmentation (UOIS), the use of large pre-trained models in image segmentation, estimating uncertainty in segmentation, and on embodied image segmentation.

\textbf{Unseen object instance segmentation (UOIS)}
UOIS for robotics aims to find an instance segmentation of objects in the foreground, typically for a tabletop scene. Recent work leverages datasets generated in simulation using a large set of objects~\cite{xie2021unseen, xiang2020learning, lu2022mean, xie2021rice}.  A difference from common panoptic, semantic, and instance segmentation scenarios is that a depth image is assumed to be available. These methods make predictions based on both intensity cues and geometric cues.  Although our goal is ultimately to obtain an object segmentation, the crucial difference is that our method estimates a factored distribution over segmentations, which is then improved by interacting with the scene, before committing to a particular segmentation hypothesis.

\textbf{Segment Anything Model (SAM)}
Recent large vision models have shown impressive results for various tasks. SAM~\cite{kirillov2023segany} is an image segmentation model that has been pre-trained on a large dataset of 11 million images. It can produce segmentation masks through point queries or box queries. Due to the flexible prompt interface and strong performance, it has been used to improve different tasks such as 3D scene segmentation~\cite{yang2023sam3d, cen2023sad} and tracking~\cite{cheng2023segment, yang2023track}. It has also been combined with other large pre-trained models such as GroundingDINO~\cite{liu2023groundingdino} to segment objects with text-prompts~\cite{ren2024grounded}. In our work we exploit the prompting interface for uncertainty estimation.

\textbf{Uncertainty Estimation in Segmentation} Many approaches to uncertainty estimation in segmentation have produced a heat map of uncertainty over pixels~\cite{lu2022mean, kendall2017uncertainty, Badrinarayanan2017segnet}.  However, the uncertainty we care about is \textit{object level} uncertainty, rather than pixel-wise uncertainty.   Some previous approaches have produced probability distributions over relatively small image patches~\cite{Hoof2014ProbabilisticSA,pajarinen2020pomdp}.  The common failure modes in modern UOIS are over- and under-segmentation of objects, therefore representing uncertainty via distributions over grouping of individual segmentation masks is more appropriate for our setting.

\textbf{Embodied Segmentation} Using robot actions to complement and enhance visual perception has a long history in robotics and is variously known as active perception, interactive perception or embodied perception; the survey by Bohg~\etal~\cite{bohg2017interactive} reviews this body of work, which includes work on interactive/embodied segmentation.

A common strategy for interactive segmentation has been to take a bottom-up approach, starting from an over-segmentation of the scene, and identifying groupings by consistency in motion~\cite{Hoof2014ProbabilisticSA,bergstrom2011scene,legoff2017agnostic}. 
An action is chosen greedily to induce motion.
There have been a number of strategies for choosing actions.  In some cases, the explicit goal is to ``singulate'' (isolate) the objects~\cite{chang2012interactive}.
Pajarinen~\etal~\cite{pajarinen2020pomdp}, on the other hand, formulate the action selection problem as a POMDP and try to pick actions that maximize long-term reward. Very recent work from Qian~\etal~\cite{qian2024riseg} also seeks to improve segmentation based on a small number of robot interactions. The action is selected heuristically based on the pixel-wise uncertainty map from MSMFormer~\cite{lu2022mean}. It differs from our focus on exploiting a representation of uncertainty obtained from prompting large pre-trained models.
Another line of work aims to use robot interaction to gather data for self-supervised training of segmentation models~\cite{lu2023self, yu2022self, Pathak2018interaction}.  This objective is in contrast with our objective of disambiguating only the current scene.

\section{Problem setting}
Our ultimate objective is to obtain an accurate interpretation of potentially highly cluttered table-top scenes, in the form of a set of partial point clouds corresponding to individual objects in the scene.  We assume that all objects in the scene are rigid and do not address the problem of revealing completely occluded objects.  

Scene segmentation is a fundamentally ambiguous problem: it may be both difficult and unnecessary to obtain a single, exactly correct interpretation.  For these reasons, we focus on constructing a distribution over segmentation hypotheses, and updating that distribution over time given new observations in which some objects have moved.

The robot embodiment consists of
a
camera that can observe the entire scene and capture registered RGB and depth images, and
 a robot arm that can reach the observed objects and make small perturbations by ``poking" the objects.
Our goal is to produce good interpretations of the scene with a minimal amount of disturbance to the objects.

The robot is assumed to be able to make precise, local contact with objects in the scene. The pushing action is determined by selecting an initial end-effector position, orientation, and motion distance.  After executing each motion, the robot retracts to a position that leaves the scene unoccluded. 

Although our objective is to maintain a distributional estimate of the segmentation state, 
in order to compare most directly with existing segmentation methods, we will evaluate our segmentation results in terms of instance segmentation metrics on 2D image masks~\cite{xie2021rice}.  
Given a hypothesized segmentation $\{s_1, \ldots, s_{N_s}\}$ where $s_i$ is a set of pixels assigned to object $i$, and a ground-truth segmentation $\{g_1, \ldots, g_{N_g}\}$, we find an assignment 
$\phi$ mapping each hypothesized segment into a ground-truth segment (or none) that maximizes the sum of the individual F-scores, and report an overall object-size normalized ({\sc osn}) precision, recall, and F-score, 

{\small 
\begin{align*}
P_n = \frac{\sum_i P_i}{N_s},\;\;
 R_n = \frac{\sum_i R_i}{N_g},\;\;
 F_n = \frac{\sum_i F_i}{\max(N_s,N_g)}
\end{align*}
}
where $F_i = 2 P_i R_i / (P_i + R_i)$, $P_i = |s_i \cap \phi(s_i)|/|s_i|$, $R_i = |s_i \cap \phi(s_i)|/|\phi(s_i)|$.
The object-size normalized metric differs from the standard P/R/F measures~\cite{xie2021unseen} in that they explicitly average the scores over {\em segments} rather than {\em pixels}.  This ensures that simply getting a few large objects correct does not overwhelm the scores of badly segmented smaller objects, which is important for manipulation problems.
We additionally wish to achieve a good segmentation result with as little disturbance to the scene as possible.  
We do not explicitly measure the amount of motion among the objects, but do measure the improvement in segmentation quality as a function of the number of actions performed.

\section{Embodied Uncertainty-Aware Segmentation}

We propose {\it \methodname{}} (\methodabbrev{}), as illustrated in Fig.~\ref{fig:illustrative_belief}. \methodabbrev{} consists of three main components, including an {\it \uncsegmethodname{}} (\uncsegmethodabbrev{}), a belief state estimator, and an action planner, operating in closed-loop interaction with the scene. The initial \rgbd{} image is processed using \uncsegmethodabbrev{}, which builds on a promptable image-segmentation model to construct a segmentation hypothesis set. This segmentation hypothesis set is used to initialize a {\em belief state}, representing a set of hypotheses about the structure of the 3D scene. Given a belief state, an action is selected and executed, and a new \rgbd{} observation is captured and used to update the belief. Finally, we generate a set of image masks corresponding to the most likely hypothesis.

\subsection{\uncsegmethodtitle{}
}
Our method, {\it \uncsegmethodname{}} (\uncsegmethodabbrev{}), provides a novel strategy for combining multiple queries to pre-trained 2D \rgb{} image-segmentation methods with some operations on the 3D point-cloud generated from a depth image, to produce a set of possible segmentation hypotheses, together with confidence estimates.

\uncsegmethodabbrev{} approaches solving the problem from two aspects: 
\begin{itemize}
    \item A ``bottom-up'' method,
    that when queried, can return masks that cover a region of interest in an image. 
    This ensures that every region in the image can be accounted for. 
    It is essential that this method have {\bf {high recall}} so that multiple queries to this method
    is likely to return most of the correct instance masks.  We refer to this method as \bottomUpPt{}. We assume it can take densely issued query points, to form an initial set of high recall masks of the whole image. We refer to this as \bottomUpIm{}.
    \item A ``top-down'' method that returns a set of image masks with {\bf high precision}. 
    These masks are very likely to correspond to correct segments, but they may not contain all the correct segments.  We refer to this method as \topDown{}.
\end{itemize}

The general strategy could use any method meeting these requirements.  In our implementation, we use the {\it Segment Anything Model} (\sam{})~\cite{kirillov2023segany}.
Given an image, it can be queried either with a pixel location or a bounding box.

We use the {\em pixel-prompted} segmentation as our \bottomUpPt{} module and its densely issued version ({\em automatic mask generation}) as our \bottomUpIm{} module. Our experiments confirm that these two do 
indeed have very high recall.

\begin{algorithm}[b]
    \begin{algorithmic}[1]
    \Require \rgb{} image $I$, depth image $D$, camera params $\theta$, text prompt $\triangleright$, overlap threshold $\gamma$ 
    \State $P := \textrm{PointCloud}(D, \theta)$
    \State $C, U := \textrm{PartitionRegions}(I, P)$
    \State $M := \text{TDHighPrecSeg}(I, \triangleright)$
    \For{$u \in U$}:
        \State $M_u = \{m \in M \mid m \cap u > \gamma\}$
        \State $H_u = \textrm{GenerateRegionHypotheses}(u,M_u)$
    \EndFor
    \State \Return $C, \{H_u \mid u \in U\}$
    \end{algorithmic}
    \caption{\uncsegmethodabbrev{}}
    \label{alg:uncos}
\end{algorithm}

We use {\sc GroundedSAM}~\cite{liu2023groundingdino, ren2024grounded}, which utilizes {\em box-prompted} segmentation with a natural language prompt, as our \topDown{} module. {\sc GroundedSAM} takes text as input, uses {\sc GroundingDINO}~\cite{liu2023groundingdino} to generate detection bounding boxes for the text, and then prompts \sam{} to generate a binary mask for each detection box.  We query {\sc GroundedSAM} with a fixed prompt ``\texttt{A rigid object.}''.  Our experiments confirm that this method does indeed have very high precision.

The overall operation of \uncsegmethodabbrev{} is described in Alg.~\ref{alg:uncos}.  The driving insight is that segmentation uncertainty is strongly region-based. In some regions of the image, the interpretation is unambiguous and there will be a single reasonable hypothesis.  However, for other regions, for example, one containing a stack of objects on the table, it is likely that the queried model will return a variety of under- and over- segmentations. However, such ambiguity is usually restricted to the local region and generally does not interact with the interpretation of a different pile of objects.

This insight leads us to {\em factor} the segmentation distribution by partitioning the image into regions and generating a distribution over hypotheses for each region.
A distribution over segmentation hypotheses for the whole image can then be constructed by taking the Cartesian product of hypothesis distributions in each region (Fig.~\ref{fig:illustrative_belief}). If the scene is constructed in a way that no such locality can be leveraged, \uncsegmethodabbrev{} will simply treat the whole scene as one region.

The algorithm begins by using bottom-up methods to partition the image into non-overlapping regions $C,U$.  The elements of set $C$ of regions are confidently considered to contain a single object. The elements of set $U$ are regions in which the segmentation is deemed to be uncertain.  Alg.~\ref{alg:seg} describes this process in detail.  The initial call to \bottomUpIm{} generates a large number of overlapping regions.  We filter out table and background using depth information, by doing plane-estimation with RANSAC. We then construct a graph with the remaining regions as nodes, with an edge between any pair of regions with a substantial overlap.  Regions with a single hypothesis are {\em verified} by calling \bottomUpPt{} seeded at multiple randomly chosen points within the region: if this process generates substantially different segmentation results then the region will not be included in the {\em confident set} $C$.  
All remaining regions are returned in the {\em uncertain set} $U$. 

After partitioning the image into disjoint confident and uncertain regions, we start to construct segmentation hypotheses for each uncertain region. 
To aid in interpreting 
\begin{algorithm}[]
\begin{algorithmic}[1]
\Require \rgb{} image $I$, point cloud $P$, Intersection-over-min (IoM) threshold $\sigma_m$, IoU threshold $\sigma_u$, num verify tests $n$
\State $M := \text{\bottomUpIm{}}(I)$
\State $M := \textrm{RemoveBackgroundRegions}(M, P)$
\State $E := \{(i, j) : |m_i \cap m_j| / \min(|m_i|,|m_j)|) > \sigma_m \}$
\State $C := \textrm{DisconnectedNodes}(M, E)$
\State $U := \textrm{ConnectedComponents}(M-C, E)$
\For{$c \in C$}
\For{$i \in \{1, \ldots, n\}$}
  \State $m_i = \text{{BUHighRecSeg}}(I, \textrm{RandomPoint}(c))$
    \If{$|m_i \cap c|/|m_i \cup c| < \sigma_u$}
    \State $C := C - \{c\}$
    \State $U := U \cup \{\{c\}\}$
    \State \kw{break}
    \EndIf
\EndFor
\EndFor
\State \Return $C, U$
\end{algorithmic}
\caption{PartitionRegions}
\label{alg:seg}
\end{algorithm}
the uncertain regions, we query \topDown{} to generate a set of candidate object masks for the whole image. We take those masks $v_1, \ldots, v_k$ that overlap with the uncertain region $u$, to be the seed masks for constructing candidate hypotheses.
Our goal in this process is to generate a set of possible partitions $H_u$ of the region $u$, seeded by these candidates. Alg.~\ref{alg:hypoth} illustrates this process: starting with each seed mask $v_i$ (and then continuing beyond that number without seeding mask if we require more hypotheses), we subtract the seed mask out of the whole region $u$, and then randomly select a point in the remaining area to query \bottomUpPt{}.  If \bottomUpPt{} returns a new mask that has a high intersection-over-union (IoU) with the unaccounted-for area, $r$,
we accept it into hypothesis $h=\{v_i, ...\}$, remove its area from $r$, and continue until we have a set $h$ of masks that nearly constitutes a partition of our target region $u$. Importantly, we also use the point cloud $P$ to determine whether the 3D volume corresponding to a suggested mask is degenerate. Suggested masks that are flat (such as labels) will be rejected.

\begin{algorithm}[t]
\begin{algorithmic}[1]
\Require Uncertain region $u$, seed masks $v_1, \ldots, v_k$, point cloud $P$, number of hypotheses to produce $N_h$, 
thresholds $\alpha, \beta$
\State $H := \{\;\}$
\For{$i \in \{1, \ldots, N_h\}$}
\State $r := \textrm{Copy}(u)$
\If{$i \leq k$}
    \State $h := \{v_i\}$;  $r := r - v_i$
\Else
    \State $h := \{\;\}$
\EndIf
\While{$|r| > \alpha$}
\State $m := \text{BUHighRecSeg}(I, \textrm{RandomPoint}(r))$
\If{$|m \cap r|/|m \cup r| > \beta$ \\ ${}$\hspace{2em} \textbf{and not} $\textrm{IsDegenerate}(m, P)$}
    \State $h := h \cup \{m\}$ ; $r := r - m$
\EndIf
\EndWhile
\State $H := H \cup \{h\}$
\EndFor
\State $EC := \textrm{EquivalenceClasses}(H)$
\State $H^* := \{(ec[0],|ec|/N_h) \mid ec \in EC\}$
\State \Return $H^*$
\end{algorithmic}
\caption{GenerateRegionHypotheses}
\label{alg:hypoth}
\end{algorithm}

Once we have generated $N_h$ complete segmentation hypotheses for this region, we check for near duplicates. 
Two hypotheses $h_i$ and $h_j$ are considered to be duplicates if 1) $h_i$ and $h_j$ have the same number of segments, and 2) the best matching between segments in $h_i$ and those in $h_j$ has an average IoU greater than a threshold.  
Using this test, we find equivalence classes of hypotheses, and return a single representative of each class;  in addition, we compute and return a ``bootstrap" confidence measure for each hypothesis class, equal to the number of elements in its class divided by the total number of samples. We use this score to determine the most likely image-segmentation hypothesis when no physical interaction evidence is available.

Finally, in Alg.~\ref{alg:uncos}, we return the confident regions $C$, and the factored hypothesis space, with a distribution over segmentation hypotheses for each uncertain region $U$.

\subsection{3D Belief representation}

Our embodied segmentation process starts with a belief state initialized with the results of \uncsegmethodabbrev{}. 
This belief could be integrated into a general goal-directed manipulation planning process, which decides whether or not to invoke information-gathering actions, depending on its given task. The planner can select actions based on the residual uncertainty, picking actions that leads to plan success under any hypothesis (e.g., deciding to push something that might be a stack of objects from the bottom, rather than picking it up from the top, for the task of cleaning up the table).  

For the purposes of testing the uncertain segmentation and belief-update process, we embed it in a loop in which the robot takes actions with the goal of reducing uncertainty in the segmentation.
It selects an action based on the hypotheses in the initial belief, executes the action on the robot, and obtains a new \rgbd{} image of the scene after the interaction. We update the belief to both track the motion of the hypothesized objects and to get a new confidence score for each hypothesis. The process repeats for several steps.
At any point in this process, we can retrieve the hypothesis with the highest confidence for evaluation against other strategies.  

Our 3D belief representation $B = (C^\plus, U^\plus)$ retains the factored structure of the 2D segmentation output, but is lifted to 3D and aggregated over time. The set $C^\plus$ now consists of a set of 3D objects $c_1^\plus, \ldots, c_n^\plus$, represented as point clouds in a global frame.  Each region $u_{(r)}^\plus \in U^\plus$ consists of a set of region hypotheses: $u_{(r)}^\plus = (h_{(r)1}^\plus, \ldots, h_{(r)n_r}^\plus)$, each of which is an interpretation of the region. For simplicity of notation, we will drop the $(r)$ from now on. It should be ranging from one to $|U^\plus|$.
Each region hypothesis $h_{j}^\plus$ consists of a set of 3D objects $(o_{j1}, \ldots, o_{jn_{j}})$.  Each object $o_{jk}$ consists of a point cloud $\eta_{jk}$ and a confidence score $s_{jk}$ indicating the likelihood that $o_{jk}$ is either a single object or part of a large whole, that is not under-segmented.

As we get additional observations, we will adjust the confidence values $s_{jk}$.  We define a score for each region hypothesis $h_{j}^\plus$ as 

{\small
\begin{equation} 
S(h_{j}^\plus)  = \frac{1}{|h_{j}^\plus|} \sum_k s_{jk} 
  - \lambda \left [|h_{j}^\plus| - \min_m |h_{m}^\plus| \right ]\;\;
\end{equation}
}which combines the average ``wholeness'' confidence of the objects in the hypothesis with a penalty for having extra objects, thus preferring the simplest hypothesis that holds the rigidity assumption.

Since the structure of the 3D belief is the same as the 2D output of  \uncsegmethodabbrev{}, we construct the initial belief by simply using the 2D masks to extract segments from the original point cloud $P$.  We initialize all $s_{jk}$ to some fixed initial value $p_0$. 
Since the hypotheses in each region are independent of those in other ones, we take the most likely hypothesis for the whole scene to be the union of $C^\plus$ and the most likely hypothesis from each uncertain region.

\subsection{Action selection} \label{subsec:action_sel}

To demonstrate the utility of the belief representation,
we use a robot to selectively poke objects in the scene using a simple greedy strategy that attempts to select a small perturbation that will maximize information gain.  We take advantage of the factored uncertainty representation to select a region of the scene that has the highest degree of uncertainty and then select the action that, when applied to that region, induces an observation distribution that is maximally discriminating among its hypotheses.

We measure the uncertainty of a region in terms of the number of high-scoring hypotheses it has: 

{\small
\begin{equation}
    \kappa(u^\plus) = |\{h_{j}^\plus \mid S(h_{j}^\plus > \delta ) \} |
\end{equation}
}

After selecting the targeted region, we need to select an informative action. For example, if the two hypotheses for a region are about whether two horizontally aligned parts are rigidly attached, then pushing along the line connecting the part centers won't be as helpful as pushing perpendicular to that. We use the physical simulation result of motions with reconstructed world hypotheses as a heuristic for the potential information gain. 

To evaluate the informativeness of an action, we construct simulated world models corresponding to all high-likelihood complete hypotheses. The worlds are constructed by taking the Cartesian product of the likely hypothesis sets for each region: 
$W = C^\plus \times \bigotimes_{(r)} u_{(r)}^\plus$.

Each world $w \in W$ consists of a set of objects defined by partial point clouds.  In order to carry out a simulation, we need to generate completions of these objects, represented as meshes.
We follow the same object reconstruction pipeline as Curtis \etal{} ~\cite{curtis2022long}: we complete the partial point cloud using a shape completion network and vertical projection, filter out any inconsistency with the current depth image, and reconstruct a concave mesh. 

Next, we sample $k$ actions, $a$, as follows: within the selected target region, we randomly sample an object among all hypotheses for the target region. We then randomly sample a pushing direction across the centroid of that hypothesized object.
Next, we simulate the effect of each action in each world, obtaining new depth images $D_{w,a}$.  
We select the action that induces most differences between the hypotheses: 

{\small
\begin{equation}
    a^* = \argmax_a \frac{1}{|W|}\sum_w |D_{w,a} - \overline{D_{*,a}}|
\end{equation}
}where 
$\overline{D_{*,a}}$ is the averaged depth for all $w$ under $a$.
Given $a^*$, we do motion planning and execute in the real world.

\subsection{Belief update}\label{subsec:beliefupdate3d}
After executing an action, we update the belief based on the robot's observation.  We cannot take advantage of dense observations during the action execution because the object is typically occluded by the robot arm.  Instead, we capture a new \rgbd{} image after the execution has terminated.

To track each hypothesis mask, we use XMem~\cite{cheng2022xmem} as a multi-object tracker for two neighboring frames. Specifically, at each time step $t$, for each hypothesized object $o_{jk}$, we initialize XMem with $I^{t-1}$ and the 2D mask of $o_{jk}$ at $t-1$. We query XMem with the new image $I^t$ and get the updated mask. 
Compared to optical-flow-based methods such as RAFT~\cite{teed2020raft}, XMem is more robust to occlusion and can handle larger movements. 

With the tracked mask, we update the object point cloud and confidence based on our rigidity assumption. We register the point cloud $\eta_{jk}^{t-1}$ to that of tracked masked area ${\eta_{jk}^{t\downarrow}}$ using RANSAC. This gives us a rigid transformation $T_{jk}^{t}$. We use the percentage of inlier points in registered point cloud as a measure of how well the point cloud motion follows the rigid assumption. This is our current time step score $s_{jk}^{t}$. 
We assume that the point clouds are sufficiently well registered so that we can just take their union as an update: 
$\eta_{jk}^{t} \leftarrow (T_{jk}^{t}\cdot \eta_{jk}^{t-1}) \cup {\eta_{jk}^{t\downarrow}}$.
The final confidence score $s_{jk}$ is the weighted average of $\{s_{jk}^l\}_{l=1,...,t}$ where the weights are determined by the displacement from $T^l$ at each step. 

\section{Evaluation}
We are interested in answering two main questions:
\begin{itemize}
    \item Does performing \uncsegmethodname{} on a single input \rgbd{} image and generating its most likely hypothesis as output result in image segmentation results that are comparable to other SOTA methods?
    \item Does the belief state initialized via \uncsegmethodname{}{} and then updated via \methodname{} provide a good basis for selecting actions for interacting with the world?
\end{itemize}
We address these two questions in the following sections.

\subsection{Segmentation from single images}

We compare \uncsegmethodabbrev{} with several methods.  The first two are state-of-the-art UOIS methods that predict a single set of object segmentation masks directly from an \rgbd{} image:
    (1) UOIS-Net-3D~\cite{xie2021unseen}
    (2) UCN~\cite{xiang2020learning}.
The next group of methods use \sam{} in some way, but do not carry out the repeated queries as in \uncsegmethodabbrev{}. 
\begin{enumerate}
    \setcounter{enumi}{2}
    \item \sam{}: returns output of the {\em automatic mask generation} query to \sam{} without further processing.
    \item \sam{}-cluster: based on the observation that \sam{} tends to over-segment objects, we construct the connectivity graph as described in Alg.~\ref{alg:seg}, and treat every connected cluster as a segmented object. 
    \item \sam{}-per-pixel-ML: assigns the highest \textit{\sam{}-conf.} mask to the pixel if multiple masks contain it~\cite{yang2023sam3d}. \textit{\sam{}-conf.} refers to the predicted confidence from the scoring head of \sam{} that it outputs with every predicted mask.
    \item {\sc GroundedSAM}: {\sc GroundedSAM} queried  with a fixed prompt ``\texttt{a rigid object}''.
\end{enumerate}    
We consider our method, \uncsegmethodabbrev{}, and several ablations:

\begin{enumerate}
    \setcounter{enumi}{6}
    \item \uncsegmethodabbrev{} {{-- {BootstrapScore}}}: returns the hypothesis from \uncsegmethodabbrev{} that has the highest average \textit{\sam{}-conf.} value, instead of the bootstrap confidence score.    
    \item \uncsegmethodabbrev{} {{-- {\sc \topDown{}}}}: uses \uncsegmethodabbrev{} without the \topDown{} masks from {\sc GroundedSAM}.
    \item \uncsegmethodabbrev{} {{-- {\sc \topDown{}} -- D}}: an ablation that further removes the depth filter for degenerate regions. 
    \item \uncsegmethodabbrev{} + UCN:  add masks from UCN~\cite{xiang2020learning} as additional \topDown{} masks.
    \item \uncsegmethodabbrev{}: our method as described in Alg.~\ref{alg:uncos}, returns the most likely hypothesis based on the bootstrap confidence score.
\end{enumerate}

These last two methods are included to illustrate the quality of the oracle best hypothesis among all hypotheses generated by \uncsegmethodabbrev{},
rather than the one \uncsegmethodabbrev{} estimated to be best. It gives an indication of the potential performance improvements we can achieve through physical interaction.
\begin{enumerate}
\setcounter{enumi}{11}
    \item Oracle \uncsegmethodabbrev{} {{-- {\sc \topDown{}}}}: The oracle best hypothesis from \uncsegmethodabbrev{} without \topDown{} masks.
    \item Oracle \uncsegmethodabbrev{}: The oracle best hypothesis from \uncsegmethodabbrev{}.
\end{enumerate}

\paragraph{Benchmark}
We compare the performance of these methods on the OCID dataset~\cite{suchi2019ocid}, which is a standard benchmark for unseen object instance segmentation.  It consists of 2390 images of tabletop scenes. Each scene contains an average of 7.5, and up to 20 objects.
We report object-size normalized scores $P_n/R_n/F_n$. We quote the results for UOIS-Net-3D~\cite{xie2021rice} and run methods 3 to 13 on the whole dataset. We rerun UCN~\cite{xiang2020learning} using the released model from the author and compute the object-size-normalized scores. 

\begin{table}
    \centering
    \begin{tabular}{c|c|c|c|c}
    \toprule
    Method & Unc-Aware & $P_n$ & $R_n$ & $F_n \uparrow$\\
    \hline     
       UOIS-Net-3D~\cite{xie2021unseen}& \XSolidBrush &86.3 &89.1&  83.6 \\       UCN~\cite{xiang2020learning} & \XSolidBrush &86.7& 90.3 &84.1 \\ 
    \hline
       \sam{} & \XSolidBrush & 29.0 & \underline{91.2} & 28.4  \\ 
       \sam{}-cluster & \XSolidBrush & 86.3 & 82.1 & 78.7 \\ 
       \sam{}-per-pixel-ML & \XSolidBrush & 80.3 & 86.4 & 76.1 \\ 
       \sc{GroundedSAM} & \XSolidBrush &  \underline{92.7}  & 73.2 & 72.6  \\ 
       \hline
       \uncsegmethodabbrev{} \scriptsize{-- {BootstrapScore}} & \CheckmarkBold & 87.5 & 88.0 & 83.2 \\ 
       \uncsegmethodabbrev{} \scriptsize{ -- {\sc \topDown{}}} & \CheckmarkBold & 88.5  &  88.3  & 84.4 \\ 
       \uncsegmethodabbrev{} \scriptsize{ -- {\sc \topDown{}} -- D} & \CheckmarkBold & 85.6  &  88.5  & 81.8 \\ 
       \uncsegmethodabbrev{} \scriptsize{ + UCN} & \CheckmarkBold & 86.7 & 90.1 & 84.3 \\ 
       \uncsegmethodabbrev{} (Ours) & \CheckmarkBold & 89.2 & 88.9 & \textbf{85.3} \\ 
       \hline 
       \hline
       {\bf Oracle} \uncsegmethodabbrev{} \scriptsize{-- {\sc \topDown{}}} & \CheckmarkBold &90.6 & 89.8 & 87.1 \\ 
       {\bf Oracle} \uncsegmethodabbrev{} & \CheckmarkBold & 91.6 & 90.5 & 88.4 \\ 
       \bottomrule
    \end{tabular}
    \caption{Comparison of all methods on object-size-normalized (OSN) precision, recall, and F-score. Unc-Aware indicates whether the method is uncertainty-aware. \uncsegmethodabbrev{} produces segmentations with highest $F_n$.}
    \label{tab:ml}
\end{table}

\paragraph{Results}
The results are shown in Table~\ref{tab:ml}. Focusing on object-size-normalized F-score ($F_n$), we observe that \uncsegmethodabbrev{} has the highest performance of the non-oracle methods, outperforming (statistically significantly) the state-of-the-art UOIS-Net-3D and UCN methods.
Methods based directly on \sam{}, without reprompting, generally perform worse. It confirms that our iterative uncertainty-aware query process helps to distill better segmentations, from the same underlying model. Comparison between \uncsegmethodabbrev{} and {{-- {BootstrapScore}}} shows the advantage of using the bootstrap confidence measure in \uncsegmethodabbrev{} to select the best hypothesis.
Removing the masks from {\sc GroundedSAM} degrades the performance, which confirms the advantage of having both \bottomUpPt{} and \topDown{} methods.  
Removing degenerate (flat) regions based on point-cloud information helps significantly, showing the advantage of leveraging depth information in our robotics domain.
Results from {\sc GroundedSAM} have the highest precision among all methods, while those from \sam{} have the highest recall (underlined). These results confirm their suitability for use as  \topDown{} and \bottomUpPt{} methods.
Additionally, we find that adding masks from UCN to the hypothesis generation process reduces performance slightly, probably because masks from UCN have lower precision than those from {\sc GroundedSAM}. 



There is a gap between the score of the actually best hypothesis and what \uncsegmethodabbrev{} believes is the best.  
The gap between these values and those of \uncsegmethodabbrev{} illustrates that there are, in at least some cases, good hypotheses that have not yet been recognized as correct, due to image ambiguity.  

\subsection{Improving segmentation through interaction}

\begin{table}
    \centering
    \begin{tabular}{l r | r r r r | r r}
    \toprule
   & Method  & 0 & 1 & 2 & 3 & {\scriptsize$\Delta$M} &{\scriptsize$\Delta$SE}\\
 \hline
& {\sc finalFrame} & 90.9 & 91.9 & 89.9 & 90.8 & -0.1 & 0.8\\
{$F$}& {\sc random} & 87.5 & 89.7 & 90.5 & 90.2 & 2.6 & 1.5\\
& \methodabbrev{} (Ours) & 87.1 & 89.0 & 92.8 & 92.9 & {\bf 5.7} & 1.7\\
\hline
& {\sc finalFrame}  & 79.5 & 80.6 & 78.9 & 79.1 & -0.4 & 1.7\\
{$F_n$}& {\sc random} & 78.5 & 81.8 &82.0 & 82.0 & 3.5 & 2.4 \\
& \methodabbrev{} (Ours) & 78.2 & 82.7 & 86.5 & 86.5 & {\bf 8.3} & 2.4\\

\bottomrule
    \end{tabular}
    \caption{Real world segmentation results:  segmentation quality initially and after each action step; final columns report the mean (M) and standard error (SE) of the changes ($\Delta$) in segmentation quality from step 0 to 3.}
    \label{tab:realworld}
\end{table}

Once \uncsegmethodabbrev{} has produced a distribution over possible segmentations, we use it to select physical interactions with the scene in order to reduce any remaining uncertainty.  We evaluate our \methodname{} (\methodabbrev{}) system in the real world with a Franka Emika robot arm. To push the object precisely, the Franka grips a stick, as shown in Fig.~\ref{fig:illustrative_belief}. We use a bidirectional RRT for motion planning and check collisions between the arm and objects using the observed point cloud. The \rgb{} and depth images are captured by a RealSense D435i camera mounted on the gripper.
The two questions we want to answer through real-world experiments are: 1) Does \uncsegmethodabbrev{} improve the efficiency of embodied segmentation; 2) Does building local memory and doing belief updates help with image segmentation. 


Our primary method, \methodabbrev{}, uses the action-selection method from Sec.~\ref{subsec:action_sel} based on a belief initialized from the \uncsegmethodabbrev{} results, and updates using the methods from Sec.~\ref{subsec:beliefupdate3d}.  For evaluation, at each time step, we compare the highest scoring hypothesis from the 3D belief state to human-labeled ground-truth masks.
We compare \methodabbrev{} with two ablations:

\begin{itemize}
\item {\sc random}: 
We retain the belief state initialization and update methods from \methodabbrev{}, but instead of selecting actions to disambiguate the most uncertain region, we randomly select a hypothesized object to interact with and randomly select a pushing direction.
Differences in performance between this method and \methodabbrev{} can be attributed to the use of the uncertainty in the belief representation to focus action selection.

\item {\sc finalFrame}: 
We use random actions, as above, but rather than maintaining a belief state and updating it after each action, we simply take the single image of the object configuration after per interaction step, apply \uncsegmethodabbrev{} to it, and return the most likely hypothesis in \uncsegmethodabbrev{} result.  Differences in performance between this method and {\sc random} can be attributed to the aggregation of observation information over time in the belief-update mechanism.  If this method reveals improved segmentation quality from the first to last frames, it can be attributed to the random motions causing physical separation between the objects, thus makes the segmentation problem easier.
\end{itemize}

\begin{figure}[h!]
    \centering
    \rotatebox[origin=r]{90}{\includegraphics[trim={0 0 0 0cm},clip,height=0.9\linewidth]{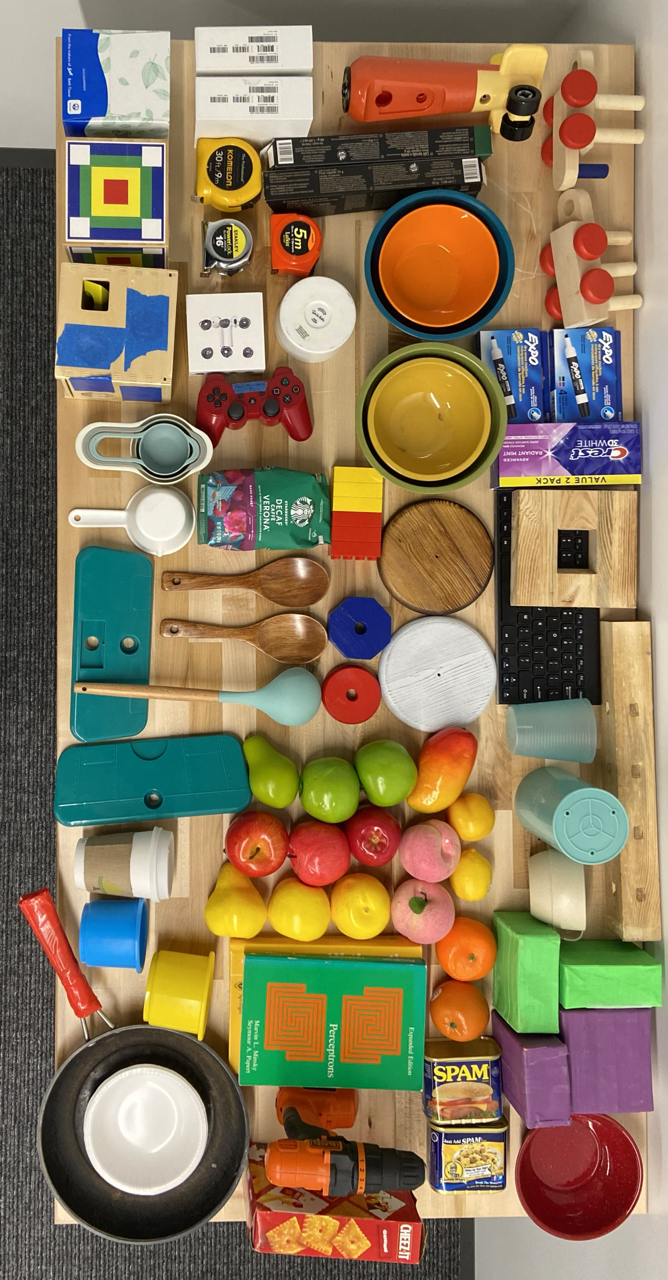}}
    \caption{Objects used for real-world evaluation.}
    \label{fig:object_collection}
\end{figure}

We set up 20 scenes with a collection of 74 diverse objects, shown in Fig.~\ref{fig:object_collection}.  We ran both the \methodabbrev{} and {\sc random} methods on each scene (the {\sc finalFrame} method uses the same images as {\sc random}, but a different method for generating a predicted segmentation).  Although the replication of the scenes for the two runs was not perfect, we set them up as similarly as possible (comparing initial images as we did so).
The robot carries out 3 actions in each scene. 

\begin{figure*}[t]
\centering
\includegraphics[width=0.985\linewidth]{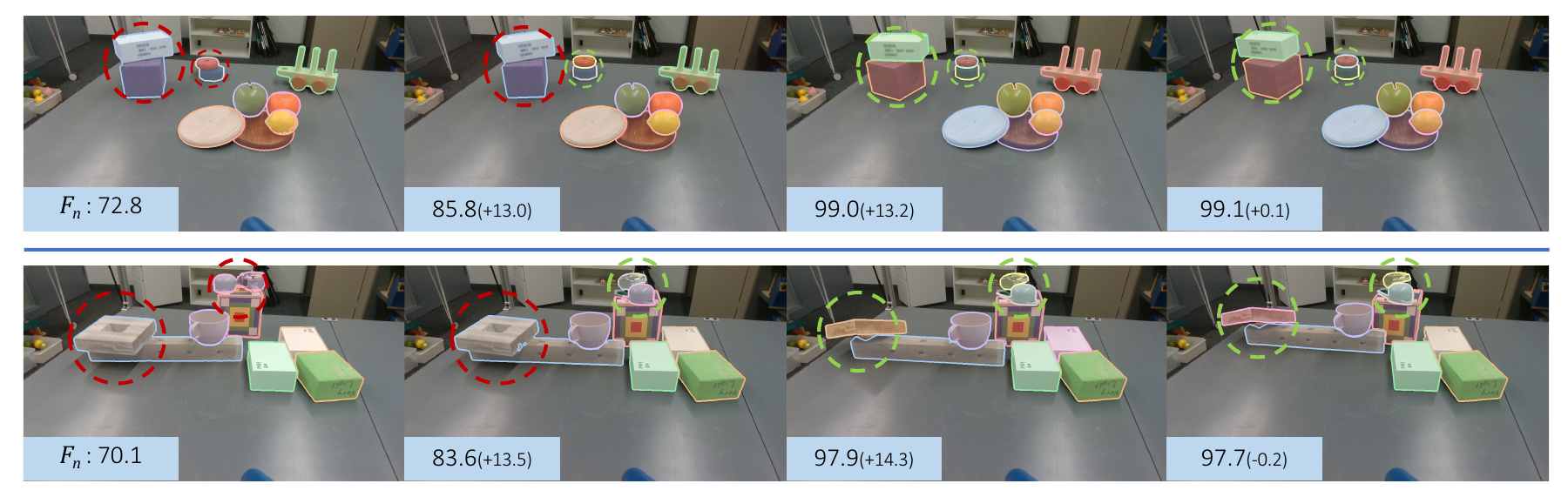}
\caption{Qualitative results from embodied segmentation. From left to right are the most likely segmentation results after 0 to 3 steps of interaction using \methodabbrev{}. Incorrect and corrected segmentations are highlighted using red and green dashed circles. $F_n$ and change in $F_n$ to the previous frame are shown in the corner.}
\label{fig:qualitative_embodied}
\vspace{-10pt}
\end{figure*}

\paragraph{Results} The average pixel-wise F-score ($F$) and object-size-normalized F-score ($F_n$),  after $K$ steps of robot interaction are listed in Table~\ref{tab:realworld}. 
Both our action selection strategy and the random strategy perform consistently better than the {\sc finalFrame} baseline. With the number of interaction steps increasing, the methods with memory get an increasing segmentation quality, and are higher than that of {\sc finalFrame}. It shows that the embodied segmentation procedure with belief update can help the robot to figure out the ambiguity in the scene and improve the segmentation quality. We include the qualitative results of \methodabbrev{} in Fig.~\ref{fig:qualitative_embodied}.

Comparing our method to the random poking baseline, there is a larger increase in segmentation quality (for both metrics) with the same number of interaction steps. This shows the benefit of having \uncsegmethodabbrev{} and belief update, which provide strong guidance for action selection in embodied segmentation.   It is also interesting to note that the {\sc finalFrame} method does not improve as a result of moving the objects, which means that the belief tracking is playing an important role in the performance of the overall system, and it is not just improving due to the physical singulation between objects.
For more results, please visit \href{https://sites.google.com/view/embodied-uncertain-seg}{https://sites.google.com/view/embodied-uncertain-seg}.

\section{Discussion}

\textbf{Limitations and Future Work.} First, our method does not utilize multi-view images to reduce the uncertainty. We are looking to incorporate active perception strategies to reduce the uncertainty caused by occlusion. Second, the current setup seeks to reduce ambiguity in the whole scene. We plan to explore a task-specific information-gathering strategy where only task-relevant regions are explored.

\textbf{Conclusion.} We formulate an \uncsegproblemname{} problem as the basis for embodied segmentation. Our method \uncsegmethodabbrev{} produces a distribution over possible segmentation hypotheses. The most likely hypothesis from \uncsegmethodabbrev{} has achieved state-of-the-art performance on the UOIS task. 
Through real-world experiments, we have demonstrated that \uncsegmethodabbrev{} can guide the embodied interaction for efficient targeted disambiguation.

\textbf{Acknowledgement}
We thank Emily Chen for helpful discussions on \sam{}.
We gratefully acknowledge support from NSF grant 2214177; from AFOSR grant FA9550-22-1-0249; from ONR MURI grant N00014-22-1-2740; from ARO grant W911NF-23-1-0034; from the MIT Quest for Intelligence; and from the Boston Dynamics Artificial Intelligence Institute.

\bibliographystyle{IEEEtran}
\bibliography{root}

\begin{thebibliography}{10}
\providecommand{\url}[1]{#1}
\csname url@rmstyle\endcsname
\providecommand{\newblock}{\relax}
\providecommand{\bibinfo}[2]{#2}
\providecommand\BIBentrySTDinterwordspacing{\spaceskip=0pt\relax}
\providecommand\BIBentryALTinterwordstretchfactor{4}
\providecommand\BIBentryALTinterwordspacing{\spaceskip=\fontdimen2\font plus
\BIBentryALTinterwordstretchfactor\fontdimen3\font minus
  \fontdimen4\font\relax}
\providecommand\BIBforeignlanguage[2]{{%
\expandafter\ifx\csname l@#1\endcsname\relax
\typeout{** WARNING: IEEEtran.bst: No hyphenation pattern has been}%
\typeout{** loaded for the language `#1'. Using the pattern for}%
\typeout{** the default language instead.}%
\else
\language=\csname l@#1\endcsname
\fi
#2}}

\bibitem{xie2021unseen}
C.~Xie, Y.~Xiang, A.~Mousavian, and D.~Fox, ``Unseen object instance
  segmentation for robotic environments,'' \emph{IEEE T-RO}, 2021.

\bibitem{xiang2020learning}
Y.~Xiang, C.~Xie, A.~Mousavian, and D.~Fox, ``Learning rgb-d feature embeddings
  for unseen object instance segmentation,'' in \emph{CoRL}, 2020.

\bibitem{lu2022mean}
Y.~Lu, Y.~Chen, N.~Ruozzi, and Y.~Xiang, ``Mean shift mask transformer for
  unseen object instance segmentation,'' \emph{ICRA}, 2022.

\bibitem{kirillov2023segany}
A.~Kirillov, E.~Mintun, N.~Ravi, H.~Mao, C.~Rolland, L.~Gustafson, T.~Xiao,
  S.~Whitehead, A.~C. Berg, W.-Y. Lo, P.~Doll{\'a}r, and R.~Girshick, ``Segment
  anything,'' \emph{arXiv:2304.02643}, 2023.

\bibitem{zhang2022dino}
H.~Zhang, F.~Li, S.~Liu, L.~Zhang, H.~Su, J.~Zhu, L.~M. Ni, and H.-Y. Shum,
  ``{DINO}: {DETR} with improved denoising anchor boxes for end-to-end object
  detection,'' \emph{ICLR}, 2023.

\bibitem{xie2021rice}
C.~Xie, A.~Mousavian, Y.~Xiang, and D.~Fox, ``Rice: Refining instance masks in
  cluttered environments with graph neural networks,'' in \emph{CoRL}, 2021.

\bibitem{yang2023sam3d}
Y.~Yang, X.~Wu, T.~He, H.~Zhao, and X.~Liu, ``{SAM3D}: Segment anything in 3d
  scenes,'' \emph{ICCV Workshop}, 2023.

\bibitem{cen2023sad}
J.~Cen, Y.~Wu, K.~Wang, X.~Li, J.~Yang, Y.~Pei, L.~Kong, Z.~Liu, and Q.~Chen,
  ``{SAD}: Segment any {RGBD},'' \emph{arXiv preprint arXiv:2305.14207}, 2023.

\bibitem{cheng2023segment}
Y.~Cheng, L.~Li, Y.~Xu, X.~Li, Z.~Yang, W.~Wang, and Y.~Yang, ``Segment and
  track anything,'' \emph{arXiv preprint arXiv:2305.06558}, 2023.

\bibitem{yang2023track}
J.~Yang, M.~Gao, Z.~Li, S.~Gao, F.~Wang, and F.~Zheng, ``Track anything:
  Segment anything meets videos,'' in \emph{arXiv preprint arXiv:2304.11968},
  2023.

\bibitem{liu2023groundingdino}
S.~Liu, Z.~Zeng, T.~Ren, F.~Li, H.~Zhang, J.~Yang, C.~Li, J.~Yang, H.~Su,
  J.~Zhu, \emph{et~al.}, ``Grounding {DINO}: Marrying {DINO} with grounded
  pre-training for open-set object detection,'' \emph{arXiv preprint
  arXiv:2303.05499}, 2023.

\bibitem{ren2024grounded}
T.~Ren, S.~Liu, A.~Zeng, J.~Lin, K.~Li, H.~Cao, J.~Chen, X.~Huang, Y.~Chen,
  F.~Yan, Z.~Zeng, H.~Zhang, F.~Li, J.~Yang, H.~Li, Q.~Jiang, and L.~Zhang,
  ``Grounded {SAM}: Assembling open-world models for diverse visual tasks,''
  \emph{arXiv preprint arXiv:2401.14159}, 2024.

\bibitem{kendall2017uncertainty}
A.~Kendall and Y.~Gal, ``What uncertainties do we need in {Bayesian} deep
  learning for computer vision?'' in \emph{NeurIPS}, 2017.

\bibitem{Badrinarayanan2017segnet}
V.~Badrinarayanan, A.~Kendall, and R.~Cipolla, ``{SegNet}: A deep convolutional
  encoder-decoder architecture for image segmentation,'' \emph{IEEE T-PAMI},
  2017.

\bibitem{Hoof2014ProbabilisticSA}
H.~V. Hoof, O.~Kroemer, and J.~Peters, ``Probabilistic segmentation and
  targeted exploration of objects in cluttered environments,'' \emph{IEEE
  T-RO}, 2014.

\bibitem{pajarinen2020pomdp}
J.~Pajarinen, J.~Lundell, and V.~Kyrki, ``{POMDP} manipulation planning under
  object composition uncertainty,'' \emph{IEEE T-RO}, 2023.

\bibitem{bohg2017interactive}
J.~Bohg, K.~Hausman, B.~Sankaran, O.~Brock, D.~Kragic, S.~Schaal, and G.~S.
  Sukhatme, ``Interactive perception: Leveraging action in perception and
  perception in action,'' \emph{IEEE T-RO}, 2017.

\bibitem{bergstrom2011scene}
N.~Bergstr{\"o}m, C.~H. Ek, M.~Bj{\"o}rkman, and D.~Kragic, ``Scene
  understanding through autonomous interactive perception,'' in \emph{ICCV},
  2011.

\bibitem{legoff2017agnostic}
L.~K. Le~Goff, G.~Mukhtar, P.-H.~L. Fur, and S.~Doncieux, ``Segmenting objects
  through an autonomous agnostic exploration conducted by a robot,'' in
  \emph{IEEE IRC}, 2017.

\bibitem{chang2012interactive}
L.~Chang, J.~R. Smith, and D.~Fox, ``Interactive singulation of objects from a
  pile,'' in \emph{ICRA}, 2012.

\bibitem{qian2024riseg}
H.~H. Qian, Y.~Lu, K.~Ren, G.~Wang, N.~Khargonkar, Y.~Xiang, and K.~Hang,
  ``{RISeg}: Robot interactive object segmentation via body frame-invariant
  features,'' \emph{ICRA}, 2024.

\bibitem{lu2023self}
Y.~Lu, N.~Khargonkar, Z.~Xu, C.~Averill, K.~Palanisamy, K.~Hang, Y.~Guo,
  N.~Ruozzi, and Y.~Xiang, ``Self-supervised unseen object instance
  segmentation via long-term robot interaction,'' \emph{RSS}, 2023.

\bibitem{yu2022self}
H.~Yu and C.~Choi, ``Self-supervised interactive object segmentation through a
  singulation-and-grasping approach,'' \emph{ECCV}, 2022.

\bibitem{Pathak2018interaction}
D.~Pathak, Y.~Shentu, D.~Chen, P.~Agrawal, T.~Darrell, S.~Levine, and J.~Malik,
  ``Learning instance segmentation by interaction,'' in \emph{CVPR Workshop},
  2018.

\bibitem{curtis2022long}
A.~Curtis, X.~Fang, L.~P. Kaelbling, T.~Lozano-P{\'e}rez, and C.~R. Garrett,
  ``Long-horizon manipulation of unknown objects via task and motion planning
  with estimated affordances,'' in \emph{ICRA}, 2022.

\bibitem{cheng2022xmem}
H.~K. Cheng and A.~G. Schwing, ``{XMem}: Long-term video object segmentation
  with an {Atkinson-Shiffrin} memory model,'' \emph{ECCV}, 2022.

\bibitem{teed2020raft}
Z.~Teed and J.~Deng, ``{RAFT}: Recurrent all-pairs field transforms for optical
  flow,'' \emph{ECCV}, 2020.

\bibitem{suchi2019ocid}
M.~Suchi, T.~Patten, D.~Fischinger, and M.~Vincze, ``Easylabel: {A}
  semi-automatic pixel-wise object annotation tool for creating robotic {RGB-D}
  datasets,'' in \emph{ICRA}, 2019.

\end{thebibliography}
\end{document}